# Intelligence and Motion Models of Continuum Robots: an Overview

**O. Shamilyan[1,2], I. Kabin[1], Z. Dyka[1,2], O. Sudakov[3,4], A. Cherninskyi[4], M. Brzozowski[1] and P. Langendoerfer[1,2]**

[1]IHP – Leibniz-Institut für innovative Mikroelektronik, Frankfurt (Oder), 15236 Germany
[2]BTU Cottbus-Senftenberg, Cottbus, 03046 Germany
[3]Medical Radiophysics Department Taras Shevchenko National University of Kyiv, Kyiv, 01601 Ukraine
[4]Department of Cellular Membranology Bogomoletz Institute of Physiology, Kyiv, 01024 Ukraine

Corresponding author: O. Shamilyan (e-mail: shamilyan@ihp-microelectronics.com).

**ABSTRACT** Many technical solutions are bio-inspired. Octopus-inspired robotic arms belong to continuum robots which are used in minimally invasive surgery or for technical system restoration in areas difficult-to-access. Continuum robot missions are bounded with their motions, whereby the motion of the robots is controlled by humans via wireless communication. In case of a lost connection, robot autonomy is required. Distributed control and distributed decision-making mechanisms based on artificial intelligence approaches can be a promising solution to achieve autonomy of technical systems and to increase their resilience. However these methods are not well investigated yet. Octopuses are the living example of natural distributed intelligence but their learning and decision-making mechanisms are also not fully investigated and understood yet. Our major interest is investigating mechanisms of Distributed Artificial Intelligence as a basis for improving resilience of complex systems. We decided to use a physical continuum robot prototype that is able to perform some basic movements for our research. The idea is to research how a technical system can be empowered to combine movements into sequences of motions by itself. For the experimental investigations a suitable physical prototype has to be selected, its motion control has to be implemented and automated. In this paper, we give an overview combining different fields of research, such as Distributed Artificial Intelligence and continuum robots based on 98 publications. We provide a detailed description of the basic motion control models of continuum robots based on the literature reviewed, discuss different aspects of autonomy and give an overview of physical prototypes of continuum robots.

## I. INTRODUCTION

Recently we are witnessing a transformation from machines that merely repeat simple action in a high frequency towards machines that solve complex tasks more and more autonomously. This change is notable in autonomous driving as well as in factory automation. Resilience is an essential concern in these fields as humans' lives are exposed to risks caused by malfunctioning devices. The same fact holds true for cyber-physical systems of systems (CPSoS) as they are used in application areas such as e-health and are the parts of automation systems. All the mentioned systems especially CPSoS should be capable of adapting themselves and their behaviour to fluctuation of environmental parameters to internal and external faults and to malicious manipulations.

They should ensure a certain level of quality of service under any conditions. If the described capabilities become real then such a system would be a resilient one.

Implementing resilient systems is a complex task. Fig. 1 shows the core features of resilience and main means allowing their implementation. Different cryptographic protocols and algorithms can be used for implementing data protection and for realizing security goals. Redundancy is a usual means to reach reliability. Autonomy can be reached by Artificial Intelligence (AI). Here we define autonomy as an intelligent and fast reaction of a CPS on dynamically changing environmental parameters and/or inner states of CPSs including the ability to find a solution of a problem by the CPS itself without human interaction.

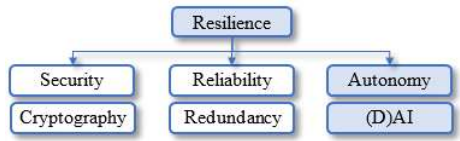

FIGURE 1. Resilience as a set of features and means for their implementation.

This definition assumes a much more complex behaviour than a pre-programmed straightforward functioning without human intervention, i.e. systems have to be able to make relevant decisions by themselves. To understand which algorithms are needed for implementing autonomy as a feature of resilient systems, it is necessary to know what are the main tasks of self-organised and self-learning systems and how these systems can detect a problem, make decisions and select one of (many) possible ways to solve a problem.

Many technical solutions are based on the properties of biological systems. The decision-making mechanisms of biological systems are often based on their learned experience. The set of learned situations can significantly influence the criteria for selecting a suitable solution during a decision-making process. In our previous papers [1], [2] we discussed octopuses as biological systems that are illustrative examples of natural Distributed Intelligence.

The Octopus body is remarkable not only from a biological perspective but also from a technical one. One of the issues is, that biologist researchers do not understand how exactly the distributed intelligence is organized. That is why it is not possible to replicate the behaviour of octopus brains. Currently, biologists are trying to understand the octopuses as living systems on one hand. On the other hand, AI developers are trying to create algorithms that simulate octopus behaviour without a biological basis. A thorough understanding of how octopuses perceive information from their environment, process it and react to it can be very helpful knowledge for understanding and implementing of distributed AI methods. In order to investigate distributed AI methods we used the Octopus arm as a suitable example. Octopus-inspired robotic arms belong to continuum robots. Unlike the usual rigid discrete robots, they possess higher flexibility, dexterity, infinite degrees of freedom and lower weight. These properties are achieved mainly due to the absence of rigid links. Due to their flexible design, continuum robots can be used in minimally invasive surgery and can reach difficult-to-access areas. In minimally invasive surgery as well as in operations to reach difficult-to-access areas it is important to have proper control of robot motions to omit additional damages and complications. That is the reason why researchers now are focused on manual robot control and path calculations in these fields. Here we see challenges related to remote robot control i.e. latency issues in IP-based networks that require the robot arm to make at least some decisions on its own. Thus we envision that in the future robots controlled by AI can rely on sensor data and react correctly to what they "see". It is even conceivable that individual parts of a robot can make decisions and solve problems by themselves locally without any communication with their control centre. This principle is similar to the hypothetical mechanism of the octopuses' neural system operation, which assumes that central brain does not fully control the arms and the arms can make some independent decisions. To allow our vision to become true we see an essential need to combine different fields of research such as DAI and robotics and to implement a working prototype for running suitable experiments. Thus the main topics and contributions of this survey paper are as follows:

- Classification of DAI field based on 30 published papers. We described Multi-Agent System as a subfield of DAI and reasoned, why this technique is promising for practical implementation of DAI.
- Overview of continuum robot types using examples of prototypes and models developed by continuum robot laboratories.
- Overview and comparison of basic types of motion control models for continuum robots and discussion of their advantages and disadvantages.
- Overview of experimental prototypes as a basis for evaluation of the applicability of the physical robot prototypes for motion control modelling as well as for the investigation of DAI methods in our future work.

The rest of this paper is structured as follows. In section II we provide information about distributed artificial intelligence and its biological roots. Section III describes common types of continuum robots and selected laboratories and projects concentrating on investigations of a certain type of continuum robots. In section IV we explain basic motion control models, discuss their advantages and disadvantages, and provide information about model extensions. In section V we discuss the intelligence as the main means to reach autonomy of continuum robots. In section VI we provide information about experimental prototypes which we found in the literature. Additionally, we share our experiences in building our own prototypes and describe problems we faced and their solutions. This paper ends with a short conclusion.

## II. DISTRIBUTED INTELLIGENCE

Generally, decision-making by technical systems can be determined as a selection of a single suitable solution from a set of possible solutions using pre-defined criteria. If the set of possible solutions is big, or the application of the criteria is complex and requires a long time, an alternative decision-making mechanism is required, for example, an AI based approach.

Investigation of the Artificial Intelligence field started around the 1940s-50s along with the development of the first computers [3]. The main research subject was behaviour of a single entity (agent) during solving problems. However, the interest in AI distribution appeared in the next few decades. As a result, Distributed Artificial Intelligence (DAI) was identified as a subfield of AI [4].

DAI like many other complex technologies finds its roots in nature. Many living systems have extremely complex mechanisms of learning, which are not yet fully investigated and understood. At system level, these mechanisms are related to high-priority functions such as memory, logical thinking, emotions, intuition, and so on. Biological systems have mechanisms to observe their inner state and environmental parameters using different body sensors. Biological systems can distinguish lived situations and classify them into safe and dangerous. Biological systems can observe and compare these situations and learn as a result.

In this section, we describe octopuses as a living example of Distributed Intelligence, report on their fascinating abilities and how octopuses can become helpful in DAI development. The second part of this section contains information about DAI, its classification and its application areas.

### A. OCTOPUSES. A LIVING EXAMPLE OF DISTRIBUTED INTELLIGENCE

Octopuses are one of the most popular examples of different technical bio-inspired projects, due to their unique morphology, features of their soft body structure lacking any rigid parts [5] with a bilaterally symmetric form, eight arms, and evolved cognition, which make them one of the most intelligent of all invertebrates [5], [6]. Some authors speculate the existence of consciousness in these animals [7].

On average octopuses live about 12-18 months. Females usually die soon after hatching their eggs [6], [8]. Young individuals do not have parents' care and always scatter right after hatching. Many octopuses spend their life solitary and limit contact with other octopuses. They are described as non-social and solitary animals. This means that the individuals are gathering their fascinating skills almost on themselves which indicates even more cognition as if they would be taught from older conspecifics and thus learn from adult octopuses.

Octopuses are predators, but they can become prey as well. The soft body makes octopuses easy prey for predators, i.e. the octopuses are not robust by nature. Octopuses have to develop their own strategy to survive using their different natural features including passive and active countermeasures. An example of passive countermeasures is the body mimicry of octopuses. An example of active countermeasures is the capability of the octopuses to shoot black ink with a water jet creating a dark non-transparent, olfactory disturbing cloud allowing them to escape [9]. Octopuses are smart enough to use different tools as protection, for example, cocoanut shells and stones [9], [10]. Also, they are able to learn from the observation of other's behaviour [8], [11]-[14]. The diversity of their prey types requires different strategies for hunting or getting food in the laboratory [15]. Such ecological plasticity substantially shapes the octopus's brain making it different from the brains of other invertebrates and vertebrates as well [16].

To understand the inner structure of the brain several studies were conducted by researchers over the last 150 years. Many research works are studying how octopuses interact with the environment, make decisions, etc. The following four main components can be distinguished here: sensory perception, perception-based cognition, learning, and memory. Each of these groups has a significant impact on the complex octopus cognition [17]. Researches show the octopus' CNS supports an acute and sensitive vision system, good spatial memory, decision-making, and camouflage behaviour [8], but it consists of fewer neurons than the PNS. The PNS in the octopus' arms accounts for almost two third of all neurons.

Scientists use neurophysiological methods and modern imaging techniques for a better understanding of neurological procedures and the connection of the different brain sections [18]-[22]. A recently developed technique allows the recording of the brain's electrical activity from implanted electrodes in freely moving octopuses [23].

The octopus neural system has the highest brain-to-body mass ratio among all invertebrates and is less centralized compared to vertebrates. The whole nervous system consists of two major divisions: the central nervous system (CNS) and the peripheral nervous (PNS) system [5], [9], [24], [25]. The total number of neural cells is about 500 million, which is comparable to the brain of a dog [7]. The vast majority of neural cells are in the PNS (~ 350 million). The brain has subdivisions and consists of more than 30 lobes. It is impossible to find the direct homology in the octopus's and vertebrate's brains [26].

The central brain acts like a decision-making mechanism: it initializes and sends the neural impulse to the peripheral. However, the brain does not issue top-down commands for every small motion of the arms [27], a lot of decisions arms make by themselves. Each arm has an autonomous structure and can act alone or in coordination with other arms without coordination or control from the CNS, at least for some kinds of motions or tasks. Additional important facts are the significantly expanded sensory capabilities of the octopus's arms – they collect chemical, tactile, mechanical, proprioceptive [7] information, and probably are even light-sensitive [28]. All this helps the arms of octopuses be more or less autonomous and work independently from the central brain. The behaviour of octopuses is substantially based on the visual information that comes from well-developed eyes. Two big (120-180 million neurons [7]) optic lobes process this information making a visual system more isolated, and less integrated into a central brain compared to that of vertebrates.

Overall, the anatomical and functional organization of the nervous system of octopuses is essentially different than those of well-studied nervous systems of vertebrates, especially mammals. At the same time, it provides complex, diverse, and flexible behaviours. Thus, studying how the brain of octopuses works may help our understanding of the organization of high-order processes in the nervous system, and this may benefit in the implementation of such knowledge into computer-based technologies.

*B. DISTRIBUTED ARTIFICIAL INTELLIGENCE CLASSIFICATION*

DAI can be classified into three groups: Parallel AI, Multi-Agent Systems (MAS) and Distributed Problem Solving [29]. Parallel AI includes development of parallel algorithms and systems' designs. MAS is described as a distributed system with two or more intelligent agents. Intelligent agents are a special form of distributed computing entities (software, hardware, etc.) that are characterized by an isolated internal state, autonomy in performing their internal operations with respect to external influence and communication with other agents much as distributed objects do. In contrast to objects, the agents have a property of reactivity i.e. they can interact with the environment (using sensors, interfaces, actuators, etc.) and change their state and behaviour while the environment changes. Also, agents have a property of proactivity i.e. they perform collaborative tasks solving to reach a common goal. Distributed Problem Solving is a subfield of MAS. The purpose of Distributed Problem Solving is to represent tasks as a set of subtasks and to assign the subtasks to the agents [2], [29]. Parallel AI, MAS and Distributed Problem Solving are not mutually exclusive groups of DAI. They all can be either implemented in one system or used independently of each other.

Further in this section, we concentrate only on the MASs: their architecture, main features, requirements, benefits and limitations. This field most fully covers the interests of our research that is why we omit an extended description of Parallel AI and Distributed Problem Solving. MAS can be used in different areas: smart grids, computer networks, complex systems modelling, city environment and robotics. We are interested in the robotics implementation of MAS and especially in bio-inspired robotics. Fig. 2 represents a classification of DAI and MAS schematically. The blocks highlighted in blue show the direction of our research.

MAS represents a set of intelligent nodes – agents – which communicate with each other and with the environment they are placed in. MAS architecture is distributed and consists of two or more agents, see Fig. 3a. Each agent is an entity, which is located in the environment and makes decisions based on the data it receives from sensors. Each agent has its task and all decisions it makes are aimed to complete it. Agents make decisions and perform actions that influence and/or change their environment with the goal to complete their tasks. More information about tasks allocation can be found in [30], [31].

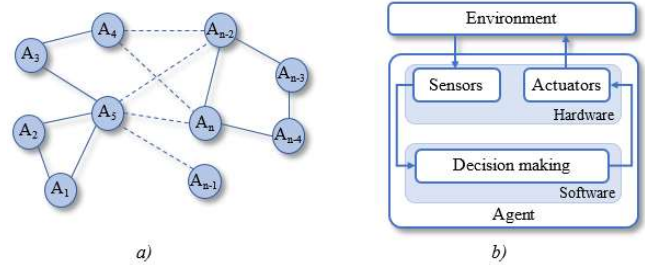

FIGURE 3. Multi-Agent System. a) Agents' organisation and their connections. Each circle represents an agent ($A_1$ – Agent 1). b) Agent's internal architecture.

Agents' internal architecture is schematically shown in Fig. 3b. Agents communicate with each other and share their knowledge about the environment. Agents often possess information only about the area, where they are located (local knowledge), that is why communication is an important part of MAS work, because it keeps a system in an up-to-date state [32].

All actions performed by the agent – cooperating and communicating with other agents, sharing knowledge about the environment and affecting it – are aimed at solving the assigned task. In a decentralized approach, each agent of MAS solves its task independently, without waiting for central control commands.

The distributed structure of MAS makes these systems flexible and scalable. Agents can be added or removed without damaging the whole system. MASs continue to work (correctly), even though some of the agents are out of service, which is the core feature of resilient systems.

According to [29] MAS has the following features:
- Leadership (leaderless or leader-follow),
- Decision function (linear and non-linear),
- Heterogeneity (homogeneous and heterogeneous),
- Agreement parameters (first, second, or higher order),
- Delay consideration (delay or without delay),
- Topology (static or dynamic),
- Data transmission frequency (time-triggered or event-triggered),
- Mobility (static or mobile agents).

This list can be used as a list of metrics, to compare the functionalities of different MAS.

In the next section, we provide a classification of continuum robots and an overview of laboratories concentrating on continuum robots.

### III. COMMON TYPES OF CONTINUUM ROBOTS

Continuum robots are recently classified into three main types: soft robots, concentric tube robots and tendon-driven robots (see Fig. 4).

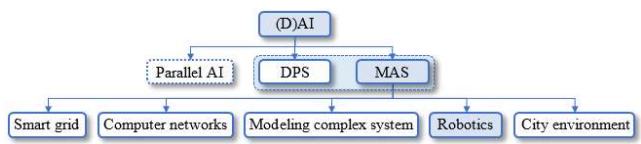

FIGURE 2. Classification of DAI and MAS.

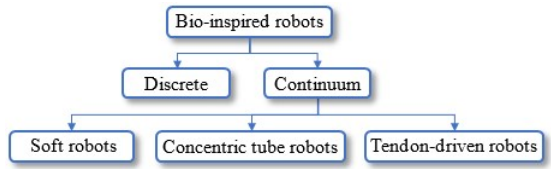

FIGURE 4. Bio-inspired robot types.

### A. SOFT ROBOTS

Soft robots are almost completely implemented using soft materials, like silicone rubber and have an infinite number of degrees of freedom. Shapes of such robots are usually inspired by worms, snakes, octopus arms, caterpillars, elephant trunks etc. Soft robotics is a very promising field, mostly because soft robots can be very useful in minimally invasive surgery [33]. One of the main challenges in soft robotic development is the actuation problem. The actuation system should be flexible and lightweight and not overload the soft robot construction.

### B. CONCENTRIC TUBE ROBOTS

Concentric tube robots consist of several elastic precurved tubes inserted one into another (see Fig. 5). Each inner tube has a smaller diameter, than the tube, in which it is inserted. Tubes usually are made of shape memory alloy. Tubes are actuated by translation and rotation motions from the tube base. The base is usually stiff. Advantages of concentric tube robots are high flexibility and small diameter (< 2 mm) [34], [35]. Concentric tube robots are mostly used in surgeries [36], for example, in neurosurgery for laser-induced thermotherapy [37].

### C. TENDON-DRIVEN CONTINUUM ROBOTS

Tendon-driven continuum robots have a different design (see Fig. 6). They usually have one central flexible backbone passing through round discs in the middle and tendons passing through discs edges (see Fig. 6a) which is one of the most common designs of tendon-driven robots. The robot can consist of several bending sections. In such multi-section robots, each section is usually controlled separately. The number and the length of each section define the length of the robot and the variety of shapes, in which the robot can be bent.

Fig. 6b shows another type of tendon-driven robot that has a larger diameter of a central backbone and does not have any disks. It is a single continuous piece with inner lumens for tendons [38]. Regardless of the type, tendon-driven manipulators are controlled extrinsically by tendon tension.

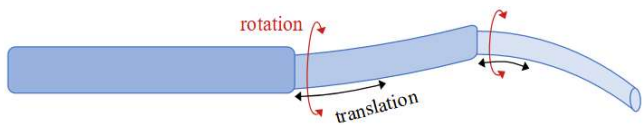

FIGURE 5. Concentric tubes continuum robot design.

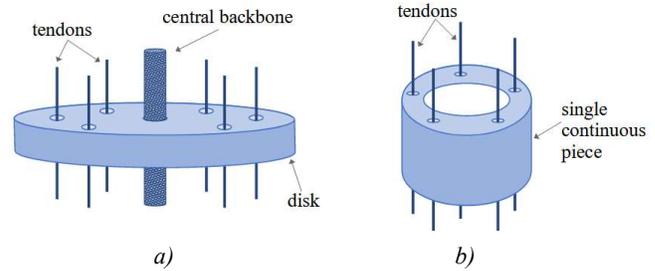

FIGURE 6. Examples of tendon-driven continuum robot design.

### D. CONTINUUM ROBOTS EXAMPLES

In the literature we found both: independent researchers [39]-[43] and research laboratories [44]-[47] specialised in the investigations of a certain type of continuum robots. The laboratories provided significant progress in their field of study. Here are some examples of their fields of activity:
- tendon-driven continuum robots, see for examples models and prototypes developed by the Continuum Robotics Laboratory in Canada [45] described in [38], [48]-[52],
- concentric tubes robots, see for examples the prototype [53] developed by ETH Zurich University [54];
- soft robotic arms:
  - driven by cables, see for example researches of the Soft Robotics Laboratory from Italy [44];
  - using pneumatic actuators as developed in the Harvard University Octobot project [47], [55];
  - exploiting hydraulic muscle actuator [51] as proposed by the Robotics and Medical Engineering Lab [46].

Table I provides examples of researches concentrating on a certain type of continuum robots.

The Continuum Robotics Laboratory investigates and develops continuum robots, and investigates ways of their control as well as human-robot interaction. Most of the laboratory researches are focused on minimally invasive surgery [45] as an application field. Mostly the laboratory works with two types of robots: tendon-driven continuum robots [38], [48], [56]-[58] and concentric tube continuum robots [36], [37]. The Continuum Robotics Laboratory developed a novel type of continuum robot [48], in which a concentric tube design and a tendon-driven design are combined. The backbone was made as a telescopic tube. Spacer disks are freely floating and equally distributed along any backbone length, which can be changed during the operation. Equal distribution is possible due to the magnet elements each disk contains. Magnets are arranged with alternating magnetic pole orientation. As a result, magnetic repulsion forces allow to keep disks at an equal distance. Tendons pass through the edges of the discs. Due to its small diameter and high flexibility, the developed robot can easily get access to different hard-accessible areas. This robot inherits the advantages of both design types. It has variable curvature per section achieved due to a telescopic backbone, which can easily change the section length. Tendon-driven mechanisms increase reachability and adjustable section stiffness [48]. A disadvantage of this prototype is that it cannot work in the environment with ferromagnetic components, due to its magnetic spacer disks.

The ETH Zurich University implemented a magnetic concentric tube continuum robot [53]. It was developed to perform minimally invasive surgery. The robot consists of two pre-curved tubes and one straight inner tube with a magnetic element fixed at the tip. Due to its design, the robot can follow the set path with different curvatures with a minimal path error. It was achieved by so-called magnetic actuation when the orientation of the magnetic field is adjusted during the robot's motion. The control of the magnetic field is provided by a special setup described in [53]. The robot can perform rotation and translation motions as well as bending deformation. The motions of the robot were validated by, computer simulations. Here the paths, which the robot follows, the robot orientation, the robot stability and the influence of the magnetic field were evaluated.

The research topics of Soft Robotics Laboratory are focused on octopuses. Many of their papers describe a development process of a soft robotic octopus-inspired arm [49], [59]-[61] as well as the experiments with control strategies [62]-[65]. Papers [50], [62]-[64] describe the development of a model-free approach, which is trained to control the motions of the octopus-inspired soft robotic arm. The model-free approach will be discussed in section IV.C. Two model representations were developed: an artificial soft robotic octopus arm and a computer model of the octopus arm. The artificial soft arm is composed of a single conical piece of silicone actuated by cables embedded inside the robot body. The prototype is morphologically similar to the octopus arm and is aimed to repeat the octopus arm's behaviour. It has a similar shape, geometric proportion and density and embedded cables act like longitudinal muscles. The fabrication process of the artificial prototype and its comparison with a real octopus arm was carried out earlier by the same group of researchers [66]. Twelve inner cables are anchored at three different distances from the robot's base along the prototype. Cables are joined into groups of four and segment the soft arm into four sections. This design gives an opportunity to control each section separately and achieve more smooth and controllable motions. All tests with a soft arm were conducted underwater. The prototype is attached to a special supporting plate. Cables are attached to four servomotors, one motor for each section. These servomotors are driven by a microcontroller, which controls the cables' tension and arm motions [49]. The Soft Robotics Laboratory also presented a computer simulation model run in MATLAB. To configure this model the Cosserat model describing deformations of physical objects was used as the approach with the most precise results. The Cosserat model will be discussed in section IV.A of this paper.

The Robotics and Medical Engineering (RoME) Lab [51] is focused on the development of bioinspired continuum robots, for example, octopus- [67] and snake-inspired [68]. The application areas can be different: healthcare, minimally invasive surgery, disaster response and area exploration. Soft, rigid and computer models are developed and presented by this lab. There are solutions with the concentric tubes design [69], [70], and tendon-driven design as well [71]. One of the developed solutions is a 3D computer simulation model based on an octopus-inspired arm prototype [51], [52], [72]. Its appearance is similar to the one of an octopus arm and is intended for underwater operations. This model is capable to implement common octopus arm deformations such as bending, elongation and contraction. The two last deformations are the common types of stretching. All motions are validated during 3D computer simulations. The model shows satisfying results and can implement bending, elongation and contraction deformations, positioning, following the path with different orientations, inspection of an object with different shapes as well as object grasping and handling [52].

Many researchers, including the 3 continuum robot laboratories listed in Table I, use the Cosserat theory for modelling motion control, which considers different kinds of deformations of physical objects, but requires a long time and significant computation costs. Many of the developed continuum robots are based on the Piecewise Constant Curvature model [73]-[77], which is a simple one but results into high errors by practical evaluation. We describe basic motion control models in the next section.

## IV. MOTION CONTROL MODELS FOR CONTINUUM ROBOTS

Flexibility of continuum robots makes them on the one hand more robust increasing their resilience, but on the other hand it makes them harder to control [78]. Different mathematical models are used to steer continuum robots. These models describe the motions of robots and define their accuracy and dexterous. Understanding of different mathematical models, used for the description of robot motions, can provide useful information for developers. For example, it can help to select the best suitable model for describing complex motion kinds. Some models allow to consider environmental influence. For example, developers can take different environmental forces into account to obtain more precise and stable motions. The models can be modified or improved according to the requirements. The high flexibility of continuum robots brings difficulties in motion control tasks. Usual control strategies, which are used for rigid robots, cannot be used for continuum robots. Therefore, new motion control strategies should be developed and used to achieve the control of continuum robots' motion.

TABLE I
EXAMPLES OF RESEARCHES CONCENTRATING ON CONTINUUM ROBOTS

| Type of robot | Type of actuation | Motion control model | Simulation/ prototype | Implemented motions and deformations | Type of control | Field(s) of application | Developed/ investigated by |
|---|---|---|---|---|---|---|---|
| Tendon-driven | Electrically driven tendons | Cosserat | Simulation and prototype | Bending, elongation, contraction | Manual control | Minimally invasive surgery | Continuum Robotics Laboratory |
| Concentric tube | Magnetic actuation | Cosserat | Simulation and prototype | Bending, rotation, translation | Manual control | Minimally invasive surgery | ETH Zürich University |
| Soft robots | Embedded cables, electrically driven | Cosserat | Simulation and prototype | Bending, reaching, fetching | Trained computer simulation model | Learning and practical field. Solution shows applicability of machine learning tools for soft robotic applications. | Soft Robotics Laboratory |
| | Pneumatic actuator | – | Prototype | Up and down motions of robot's arms only | Microfluidic logic | Foundation for a new generation of completely soft, autonomous robots | Harvard University: Octobot project |
| | Hydraulic muscle actuator | Shape Function-Based model | Simulation and prototype | Bending, elongation, contraction, fetching, object inspection | Trained computer simulation model | Minimally invasive surgery, inspection of objects with irregular shapes, object handling | Robotics and Medical Engineering Lab |

Basic motion models analyse relations between applied forces and resulting motions of the robots that allow to describe a motion of the whole robot and its parts in a three-dimensional space over time.

Despite the variety of motion control models, there exist only a few basic motion control models. Reasons for this variety are additional requirements, assumptions or tasks, which should be taken into account. In the rest of this section, we describe the basic models and their modifications.

*A. COSSERAT-BASED MODELS*

The simplest model describing a motion of physical objects assumes that the spatial extents of the objects are negligible, i.e. the mass of a physical object is concentrated in a single point called the centre of mass. Additionally, simulated physical objects have energy, velocity and translation momentum. The latter is defined as the product of the mass and the velocity of the object. The velocity and momentum are three-dimensional vectors in a Cartesian coordinate system. Under the influence of an external force F, which is also a three-dimensional vector, the motion of such physical object(s) can be described using the rules of classical mechanics.

A more complex model considers the size of physical objects, but assumes that the mass of an object is equally distributed in its volume, i.e. the density of the object is constant. The model assumes that such solid objects are not deformable, i.e. the distance between any two points of the object is always the same. Such objects are called rigid bodies. For this type of objects, the model describes two kinds of motions – translation and rotation – under the influence of external forces.

Models, which work with elastic objects, consider additionally the fact that objects can change their form under influence of external and/or internal forces, i.e. they describe deformable physical objects. Fig. 7 illustrates different kinds of motions and deformations.

Fig. 7a and Fig. 7b illustrate simple body motions without any deformations. Translation motion is shown in Fig. 7a. After this motion the resulting body is described by every point of the original body translated parallel during the motion. Fig. 7b illustrates rotation – a motion of the body around a fixed axis. Fig. 7c-f illustrates different kinds of deformation: stretching (includes elongation and contraction), shear deformation, bending, and twist deformation.

The first model for motions of a rod considering deformations is known as the Cosserat's rod model. It was proposed by the Cosserat brothers in 1909 [79]. The Cosserat model is nowadays one of the most known models applied for the description of the motions of continuum robots. A Cosserat's rod is represented as a long, thin and perfect flexible body, i.e. the Cosserat rod-based models consider stretching, shearing, bending and twisting as deformations, which can change the shape of the rod. There is a wide range of literature concerning Cosserat theory, however, many authors define space variables and forces differently, which makes the process of understanding the theory more complicated. We used variable notations defined by the authors of the Elastica Project [80] in our description of a Cosserat's rod, see Fig. 8.

Fig. 8a shows a Cosserat's rod without any deformations, i.e. in a reference state. The rod's central line is denoted as *s*. The points $s_1$ and $s_N$ on the line *s* correspond to the rod's borders in Cartesian coordinates. External force **F** and torque **M** influence the rod, whereby the external influences can be not constant, i.e. they can change in space and over the time *t*. Here $f(s,t)$ denotes distributions of the external force along the rod and over the time; $c(s,t)$ denotes the distribution of the external torque along the rod and over the time.

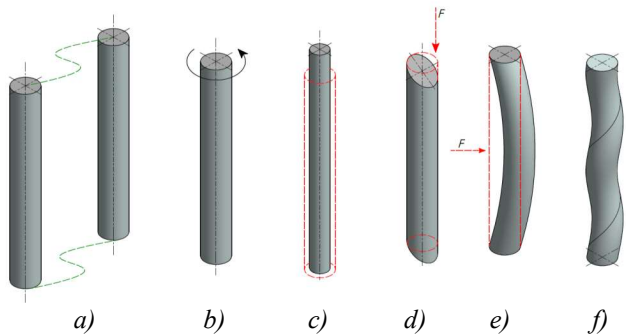

**FIGURE 7.** Different kinds of motions and deformations of a rod: a) – translation; b) – rotation; c) – stretching: d) – shearing; e) – bending; f) – twisting. On some deformations initial state is marked with a red dashed line.

Fig. 8b shows the rod in a deformed state. The vectors $i$, $j$, $k$ represent the three-dimensional Cartesian coordinates. The rod's central line $s$ is now deformed. The external influences caused reactions – the motions and deformations of the rod, as well as the rod's internal force $n(s,t)$ and torque $\tau(s,t)$, which are distributed along the deformed rod's centreline $s$ and are changing over time.

A part of the rod, for example, between points $s_i$ and $s_j$ we denoted as $\Delta s$. If the selected part is small ($\Delta s \rightarrow 0$) we denoted it as $ds$ (see Fig. 8c). It is assumed that $ds$ is a straight line.

Each point $s_i$ of the line $s$ can be described using the vector $r(s_i,t)$, starting always in the point with the coordinates (0,0,0). Coordinates of each point of the line $s$ can change over time $t$. Additionally, for each point $s_i$ three vectors $d_1$, $d_2$, $d_3$ are defined. Vectors $d_1$ and $d_2$ span a cross-section of the rod in a current point of the line $s$. Vector $d_3$ defines the normal of the cross-section and can be calculated as $d_3 = d_1 \times d_2$. The three vectors $d_1$, $d_2$, $d_3$ are called *directors*. The directors' orientation is the result of the rod deformation. In Fig. 8b $d_3$ director's orientation is co-directed with a tangent to the line $s$ in the point $s_N$. Some kinds of deformations – shear or extension – can change the direction of the $d_3$ vector. In this case, there is an angle between the $d_3$ vector and the tangent to the centreline $s$. This angle defines the so-called shear strain vector $\sigma$.

The internal force $n(s,t)$ and the torque $\tau(s,t)$ in a current point $s_i$ of the line $s$ determine the position of the point $s_i$ in space over time as well as the orientation of the directors in this point. Thus, equations for $r(s,t)$ and directions of the vectors $d_1$, $d_2$, $d_3$ in each point of the line $s$ over the time $t$ describe the motion of the rod. These equations can be used for the development of computer simulation models. An important fact is that the motions of a Cosserat's rod can be caused not only by external forces but also just by internal forces, for example, muscle contraction.

The development process of a computer simulation model of a Cosserat rod can be complex due to the sophistication of the Cosserat model itself. First of all, the equations used by Cosserat's model usually can be solved only using numerical methods. Second, the model describes a rod as a continuum object and calculates motion for each rod's point, i.e. the number of the points tends to infinity. Calculations also include the orientation of directors in each rod's point. Due to this fact, a discretization of a continuum Cosserat rod is necessary. The continuum rod is usually represented as a certain number of segments of the length $\Delta s$ (instead of $ds$). Afterwards, the rod is represented as a set of segments. The higher the number of segments, the more precise the description of the motions using the model.

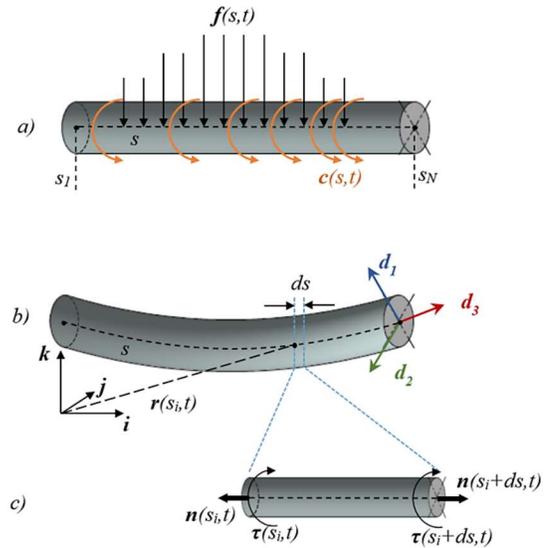

**FIGURE 8.** The Cosserat rod model: a) rod in a reference state with applied distributed external force $f(s,t)$ and torque $c(s,t)$; b) rod in a deformed state; c) segment of the rod $ds$ with internal force $n$ and torque $\tau$.

The Cosserat model has nowadays many variations, each of them focused on different specificity sets of external forces and loads and deformation types. Real forces and loads, which can affect a robot prototype, for example, the gravity force, should be taken into account. The model should be created in accordance with robot specifics (type of robot, material), workspace and set of motions. Thus, many extensions of the Cosserat theory for different types of continuum robots are developed. In the rest of this subsection, we describe some of the extensions to provide a look at a variety of model extensions and their use-cases.

In [49] authors applied the Cosserat model for the motion control of a multi-bending octopus-inspired soft robotic arm. The model takes into account external loads such as gravity, buoyancy, drag force, added mass and cable weight. The model was experimentally validated using a single-section silicone prototype that works in water. For model validation authors chose bending, deformation and reaching and fetching motions. These deformation and motions are typical for octopuses. Reaching is a spread of a local small bend from an arm base to its tip (octopus pushes its arm away to reach an object). While fetching, an octopus pulls a reached object back, for example, octopus brings food to its mouth [81], [82].

The basic Cosserat model can be applied for tendon-driven as well as for concentric-tube continuum robots. Authors of [83] consider the concentric-tube continuum robot as a set of so-called precurved tubes and applied the Cosserat model to the set. A precurved tube is not a straight one, but has already its own curvature. The authors provide equations of a single precurved tube with external loads. These results are applied to a set of tubes, i.e. to the complete robot. Each tube of the robot has its own initial configuration, i.e. its arc-length and curvature. The developed model was experimentally validated using a concentric-tube continuum robot designed by the authors [83]. During the validation, various loading conditions and actuator configurations were applied to the robot.

The Cosserat-based motion control model described in [84] was specially adjusted for tendon-driven continuum robots, in which tendons go through the whole length of the robot. The model was evaluated in [84] using their robot prototype. The body of the robot prototype consists of disks and tendons. Each tendon goes through all disks and knots at the last disk. A backbone tendon locates in the middle and other tendons are located around the backbone. The authors of [84] assumed that the body of the robot is a cylinder and can be considered a Cosserat rod. For the motion control of the robot, the Cosserat rod and Cosserat string models were coupled. The Cosserat rod model was used for the rod motion control description. A simplified Cosserat model, called Cosserat string model, describes tendons and takes into account force distribution on tendons and tendon loads on the rod. The Cosserat string model assumes that the size of the cross-section of a Cosserat rod is negligible, i.e. the area of the cross-section of a Cosserat rod tends to zero. The consequence is that not all deformations have to be taken into the account by the Cosserat string model, for example, a shearing. For verification of the robot's motions, the authors of [84] checked the tip error by applying different external loads to a self-designed tendon-driven robot that was fixed on a base.

The model described in [56] was developed as a special case of the previously described model of the tendon-driven continuum robots [84]. It was developed for a tendon-driven continuum robot with extensible sections. The robot prototype combines tendon-driven and concentric-tube continuum robot design. It has an elastic backbone with extensible sections in the middle and 3 tendons around [48]. In [56] calculations of varying section stiffness and force distribution are given. The model was validated using computer 3D simulations. The simulated robot had to move along a given path with a minimum path and tip errors.

### B. PIECEWISE CONSTANT CURVATURE MODEL

A widely used motion control model is the Piecewise Constant Curvature (PCC) model. Similar to the Cosserat string model, the rod cross-section size and the mass distribution in the rod are not taken into account by the PCC model. The PCC model is considered a simple robot motion control model due to smaller number of parameters used for the geometrical description compared to the Cosserat model. In this subsection, we give a short overview of the PCC model with an example of its application.

We explain this model using the example shown in Fig. 9. The robot shape is represented by a smooth, continuous curved line, which has no acute angles, i.e. it is not a polygonal chain. Moreover, the mass of the robot and external loads are not considered. Corresponding to the PCC model, the centerline $s$ is partitioned into a finite set of segments. Segments have different lengths. During the deformation, the segments do not elongate or contract, so the length of each segment $l_{segi}$ remains constant. PCC model assumes that the curvature $K=1/r$ of a segment (where $r$ is the radius of the curve) is constant for the segment. Thus, the PCC model associates each segment with a single circular arc, which most accurately repeats the curvature of the segment. Each segment has its angle of bending $\theta_{segi}$, which with $r_{segi}$ can be used to calculate $l_{segi}$. Our example shown in Fig. 9 has three segments: $seg_1$, $seg_2$, $seg_3$ [56]. Each segment is associated with a circular arc that is a part of a circle with a radius $r_{segi}$. The circles are marked with a blue dashed line[1] in Fig. 9 and the arc projection of $seg_1$ on the $ij$-plain is marked orange. Each circular arc is described by a set of three arc parameters: curvature $K_{segi}$, angle $\varphi_{segi}$ between an axis and the arc projection on a plane and arc length $l_{segi}$. The arc parameters and coordinates of the arc centers can change over time. The PCC model transforms the arc parameters ($K_{segi}$, $\varphi_{segi}$, $l_{segi}$) into the Cartesian coordinates ($\boldsymbol{i}$, $\boldsymbol{j}$, $\boldsymbol{k}$), and allows to calculate robot's position, i.e. the coordinates of each centreline point of the robot over time, similar to Cosserat model. A detailed description of the PCC model can be found in [85].

The PCC model describes motion of a Cosserat string of a constant length. Due to this limitation, the PCC model is inefficient in many practical cases. For example, it does not suit the cases, where external loads (for example, gravity) have to be taken into account [49] or if the mass distribution is not negligible, i.e. the PCC model is, for example, not applicable to many soft robots [86]. Despite the disadvantages describing soft robots, the PCC model is considered the simplest motion control approach that can be applied for continuum robots [65], [85] and especially for tendon-driven continuum robots.

To address the weaknesses of the PCC model for motion control of soft robots, a new approach was developed and described in [59]. The approach is called Piecewise Constant Strain. It couples Cosserat rod and PCC models and brings advantages from both of them: representation of robots segments as circular arcs, calculation not only of constant curvature and elongation, but also of torsions and shears, taking into account external loads and torques and implementation of different actuation solutions. Authors of the Piecewise Constant Strain model [59] used the PCC model to discretize a continuous model developed with the Cosserat approach. Eventually, the continuous model was described as a finite set of strain vectors. The Piecewise Constant Strain model was validated using a multi-section soft robot octopus-inspired arm in different variations of bending deformation.

---

[1] Fig. 9 is drawn using the following axis foreshortening: 0.5 for $\boldsymbol{i}$, and 1 for $\boldsymbol{j}$ and $\boldsymbol{k}$. That is the reason, why the circles are represented as ellipses.

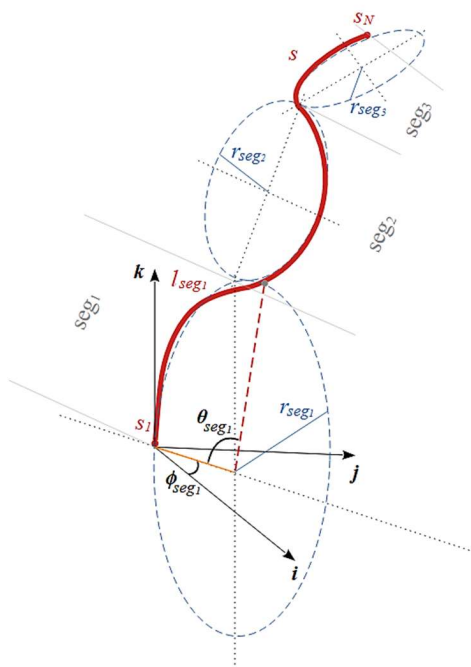

**FIGURE 9.** Piecewise Constant Curvature model.

The performance of the Piecewise Constant Strain model was compared in [59] with the Cosserat rod model using the same octopus-inspired multi-section soft robot arm for both experiments. As a result, the Piecewise Constant Strain provides faster calculations.

### C. MODEL-FREE APPROACH

A motion approach described without the application of any motion control model, is called model-free. It is a new field in developing continuum robot control strategies. This field of study is very promising and offers many possibilities for investigation and development. The survey [65] pointed out that the model-free approach shows better performance and results, even though it is represented as a simple neural network.

One of the main advantages of the model-free approach is its independency on any robot parameters such as shape, weight, length, number of joints, etc. All motion control model-based approaches, which we described above, were developed according to a robot's shape, type of actuation, type of material, environmental parameters, external loads, robot's motions, etc. The model-free approach uses Machine Learning (ML) and data from sensors, to build a control scheme with better performance. Also, the approach is well suitable for solving nonlinear and distributed problems.

An example of the model-free approach is presented in [62], [63]. These papers describe the development and application process of the model-free motion control strategy based on ML. Authors of [62], [63] used a soft robotic octopus-inspired arm for their experiments. They created a computer simulation model and collected sample data of robot's motions. The sample data was obtained by actuator babbling (random motor actuation) and consists of random environment exploration motions performed by the robot. We assume, that this type of data collecting, was chosen by the authors of the model-free approach, because their soft robotic prototype has infinite degrees of freedom, and therefore has an infinite number of possible motions, which are impossible to cover. The collected data was partitioned into training and testing sets. In their experiments, the authors used a nonlinear autoregressive network with exogenous inputs (NARX) [87], [88] that shows better performance and accuracy results than an ordinary recurrent neural network.

The performance of the model-free approach was compared with the Cosserat model. The comparison showed that the model-free approach based on ML is trained faster from scratch without any assumptions, which are usually made for model-based approaches.

The model-free approach provides a novel approach of creating motion control strategies. It does not require any complex mathematical calculations as other mathematical models do. Also, the model-free approach is scalable and does not depend on the robot design.

### D. SHAPE FUNCTION-BASED MODEL

Among the well-known examples of motion control models described above, there are models used only by their developers. For example, the Shape Function-Based model [89] was created specifically for an octopus-inspired soft robotic arm described in [52].

The robot consists of three sections. Each section has a variable length. The robot works underwater and has hydraulic muscle actuators. Shape Function-Based model takes into account all specifics of a robot design mentioned above. However, Shape Function-Based model does not calculate buoyancy and drag force, despite the fact that the robot works underwater. To the best of our knowledge, the Shape Function-Based model was used only by its developers. Practical use cases were described in [51], [52], [72].

The Shape Function-Based model was validated using 3D computer simulations. During the research, authors of [89] extended and improved their model. In [89] authors worked only with one-section of their robot, to simplify the model. 3D computer simulations of this version of the model perform bending, pure elongation and control stiffness of the hydraulic muscle actuators. Due to the section's variable length, the mass distribution was also taken into account. Authors presented [51], [52] and [72] as extensions of [89].

They improved their model, so that computer simulation can perform not only basic deformations, but it also can follow a given path with different tentacle orientations, inspect objects with different shapes as well as grasp the object and manipulate it.

### E. MOTION CONTROL MODELS: SUMMARY AND DISCUSSION

In this subsection we summarize information about the motion control models described above. Fig. 10 represents motion control models and their variations in a tree-like structure.

Blocks with the same colour indicate that these models belong to the same laboratory or research group. Yellow blocks belong to the Soft Robotics Laboratory, red blocks – Vanderbilt University, Nashville, the violet block – to the Continuum Robotics Laboratory and the green block – to the Robotics and Medical Engineering Lab.

Applying a mathematical model causes complex and time-expensive calculations. Computer simulations of the motions corresponding to a mathematical model are based on the discretisation of the mathematical model. Depending on the parameters selected for the model discretisation, computer simulations require different calculation time, whereby the quality of the motion simulations can differ significantly, for example, path or tip error can increase.

A model-free approach using an ML algorithm does not need to solve a complex system of differential equations that is required when applying a mathematical model or implementing the model as computer simulations. The advantages of model-free approaches are the low computational costs and – as consequence – the short execution time. The problem is that ML algorithms need a training set of motions for learning the motions. This training set of the motions can be obtained using computer simulations. If the ML uses a set of simulated motions, for example, a computer simulation of the motions based on the Cosserat model, many important "details" can be missed in such virtual reality. Simulated, i.e. virtual motions, deal with "ideal" Cosserat rods/strings. Real prototypes can have many minor details, which are not considered by the ideal model, for example, the mass density of the Cosserat road can be not constant in the real world, the tendon can be not a centerline, etc. Even more, the robot prototypes can (significantly) differ from each other, due to deviations in their fabrication process, i.e. their mass, length and other parameters are not the same, even if the goal was to fabricate two identical prototypes. Simulation-based approaches cannot take into account these differences. This can be the reason of different path errors of different prototypes.

Such difficulties can possibly be avoided using real physical motions of already fabricated prototypes. In this case, the same learning process applied to different prototypes can result in similar accuracy of the learned motions and reactions. Thus, the observations of the motions of a real physical prototype can be a promising alternative. Observed information can be stored, analysed and used by machine learning algorithms as a training set. Due to the advantages of the real physical motion model we decided to concentrate on it in our future experiments.

## V. INTELLIGENCE AS A MEANS FOR AUTONOMY

Practical investigations of decision-making mechanisms are extremely complex and require a prototype that is able to observe the environmental physical parameters, to control its own inner states and has controllable motion and reaction mechanisms.

In this section, we discuss the intelligence as the main means to reach autonomy of continuum robots, especially due to the fact, that the continuum robot missions are bounded with their motions/movements.

### A. DIFFERENT ASPECTS OF AUTONOMY

Autonomy of a (complex) system has many aspects:
- energy-independency;
- adaptability to fluctuation of environmental parameters and/or internal states;
- mobility (athletic intelligence);
- decision-making (cognitive intelligence).

Often energy independence is considered as the first step to reach autonomy of a system. The next important aspect is system adaptability. The adaptability can be described as a fast reaction to observed parameter fluctuations, which can be partially pre-programmed to reach a certain level of autonomy. But generally, not each situation can be pre-programmed. Autonomic robots should be able to make decisions by themselves. For the better understanding, we assume that the system:
- observes a finite set of environmental parameters $\{ep_1, …, ep_L\}$;
- observes a finite set of internal states $\{st_1, …, st_M\}$;
- has a set of pre-defined criteria $\{k_1, …, k_N\}$;
- can derive its own reaction as a pre-defined function (1) of the observed parameters and states corresponding to a criterion

$$R_{l,m,n} = f(ep_l, st_m, k_n) \qquad (1)$$

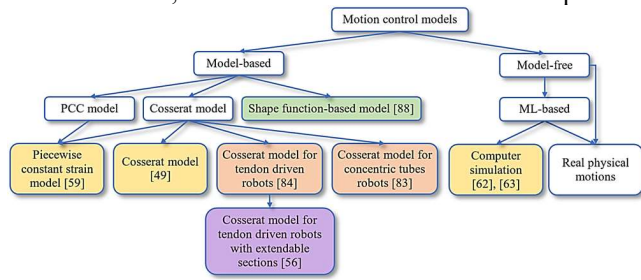

**FIGURE 10.** Dependency between motion control models for continuum robots. Blocks with the same colour represent the same laboratory/research group.

If the function $f$ is injective and only one of the environmental parameters and/or internal states is changed the corresponding reaction can be clearly selected and this selection can be programmed. For example, if a system sensitive to temperature detects that environmental temperature increases (environmental parameter $ep_1$) the system can check its inner temperature (internal state parameter $st_m$) and if it is higher than a pre-defined threshold (criterion $k_n$) a cooling system can be activated (reaction $R_{l,m,n}$). Many systems now have cooling subsystems that are activated automatically, which can be interpreted as a certain level of autonomy.

Typically, a mission of a continuum robot is to reach a difficult-to-access area and to perform a sequence of pre-defined operations, i.e. the mission is directly related to the robot's mobility. Obstacle avoidance is a situation in which a mobile robot should make a decision. The robot has to detect an obstacle and realize it, i.e. its shape, size, velocity. The obstacle's parameters, the robot's inner state (for example, the capacity of the accumulators), estimated influence of the obstacle on the success of the mission, and many other factors can define the reaction of the robot, i.e. the shape of the robot's new path. Generally, if many environmental parameters and/or internal states and/or criteria are changed, or if the function $f$ in (1) is not injective it can cause a set of possible reactions. Moreover, some reactions from the set can be contradictory. In this case, AI methods can solve the task by selecting the most appropriate reaction increasing the autonomy of the system.

Two different types of intelligence – athletic and cognitive intelligence – were initially presented by Boston Dynamics [90], [91]. They explain these terms with the example of human intelligence. Athletic intelligence copes with the body motions, such as walking, running, climbing, the way how humans control their body, hold a balance, avoid obstacles, use different ways of walking for the different kinds of landscapes. Cognitive intelligence is not about mobility, but about decision-making, planning and solving problems. By combining athletic and cognitive intelligence, Boston Dynamics achieve significant results in development of autonomic robots [92].

### B. MOTION CONTROL AS BASIS FOR INVESTIGATIONS OF ROBOT INTELLIGENCE

A computer simulation of a motion control model or a continuum robot prototype can be used as a basis for investigating decision-making mechanism. We searched for suitable open-source computer simulation software or ready to use robots or robot prototypes, which can be bought. However, we faced difficulties with finding such products. The open-source computer simulation software of a motion control model suitable for continuum robots [93] does not have sufficient description and documentation, i.e. it is not suitable for a new user. Boston Dynamics [90] leads investigations in different aspects of robot intelligence and autonomy for not continuum robots. Their robots can perform different kinds of movements and can be programmed to execute a set of movements, i.e. the so-called "athletic intelligence" is their key feature. Different kinds of "cognitive intelligence" belong to their research aspects too [91]. A well-known example of the cooperative work of two (at least partially) autonomous devices is "Spot" – an agile mobile robot [92], developed by Boston Dynamics. These robots can be purchased but their operation principles are kept secret by the developers. The researchers from ETH Zurich describe this problem its causes and consequences in [94]. Also, there is no information in the literature about how autonomy and intelligence of the robot were implemented. Most of the commands cannot be controlled by the users, so robot's behaviour can be considered as a black box.

## VI. OUR CONTINUUM ROBOT PROTOTYPE
### A. AVAILABLE CONTINUUM ROBOT PROTOTYPES

Due to the non-availability of a simulation model or a physical prototype of a continuum robot suitable for our investigations of decision-making mechanism, it is necessary to implement a model or a robot. A simulation model is always based on some simplifying assumptions and cannot consider many motion and deformation aspects, otherwise, the model becomes complex and slow. Manufactured robots can differ from the "ideal" robot in the simulations, i.e. they have similar but not exactly the same weight, length, mobility, etc. Depending on the differences, the applicability of the results obtained using simulations has to be evaluated using a physical prototype. Thus, the investigations on the decision-making mechanism using a robot prototype seem to be the preferable solution.

Our search of detailed documentation for implementing a continuum robot prototype resulted in two solutions: the soft robotics toolkit [95] and the instructions for implementing a tendon-driven continuum robot [96].

A toolkit presented in [95] contains a set of instructions and a list of materials which help to implement a soft robot prototype. The soft robot prototype is intended to be pneumatically actuated. The actuation system contains a set of tubes, pumps and supplementary electric units, which complicate the process of the prototype fabrication as well as its motion control.

A robot design published on the "Hackaday" web-page [96] contains a description of a fabrication process of a pure tendon-driven robot. It has been presented as a homemade project by Joshua Vasquez. More information about this solution can be found in [97], [98]. The robot represents an octopus arm with two sections controlled separately by two manual controllers. These sections have the same length. The robot consists of spacer disks and flexible tendons. There are different tendons for the control of each tentacle's section that allow to achieve individual control of each section. A very important feature of this robot design is the so-called "back-drivability", which means the following: it is possible to manipulate the tentacle by hands, and those motions will be transmitted back through the cables to then drive the controller. In other words, it gives an opportunity not only to move the tentacle by moving controllers, but also to move the manual controllers by changing the shape of tentacle's sections.

### B. OUR EXPERIENCE AND FUTURE WORK

Despite our exhaustive search of the literature, we did not find information about modelling continuum robot motions using a physical robot prototype. We found only the sources [95], [96] describing the robot prototype implementation process. In our previous paper [1] we discussed the possibility to use the soft robot prototype described above, but the difficulties in its implementation, actuation and control would have drawn us apart from our main goal – the investigation of distributed control and distributed decision-making mechanisms. For the same reason, we have abandoned the idea of implementing the prototype [95] with pneumatic actuation. We selected the tendon-driven design as a good example of a robot with a detailed description that can be easily built and controlled. Especially the back-drivability feature seems very promising for motion learning.

We implemented our first prototype corresponding to [96] but used other materials. The initial description suggests using plastic spacer disks, but due to their fragility, we used aluminium ones [2]. Aluminium was chosen as a light-weight and robust material. Usage of other light-weight materials such as carbon fiber is also possible. Our prototype is made of 15 disks (30 mm diameter and 1 mm height), one central flexible backbone tendon (3 mm diameter) and 8 side tendons (1 mm diameter). Disks are rigidly fixed on the central backbone tendon at a distance of 20 mm from each other. As the initial prototype [96], our prototype has two sections. Side tendons pass through each disk along both sections (4 side tendons for each section). A video displaying robots construction and moving can be found in the additional materials for this paper.

Our first step to achieving autonomy was to replace manual controllers with motors that make the motion control programmable and allow to control the prototype without human-robot interaction. For motion control of our prototype we will implement model-free approach, using an exhaustive search – i.e. brute-forcing – all possible positions of a single segment as a mean to collect data about robot movements. To collect data we will move only the upper segment, while the lower segment will be rigidly fixed in a vertical straight position. Obviously, the collected dataset will contain some overlaps, where there is more than one way to reach the same point of robot's workspace. These overlaps are especially interesting for us in the investigation of decision-making mechanisms, i.e. how the robot will choose the way to reach the point if there are multiple variants.

### VII. CONCLUSION

Recently, resilience became a very important feature, due to the growing demand for autonomous CPSoS. Such systems should be self-aware and have the ability to adapt their behaviour in accordance with a dynamically changing environment. Distributed control and distributed decision-making mechanisms can be a suitable solution to achieve autonomy in CPSoS, however, these methods are not well investigated, yet.

In this paper, we gave an overview of DAI with detailed description of MAS, and a description of the living example of distributed intelligence – octopuses. We gave an overview of continuum robots state-of-the-art, their common types and designs, as we believe that a continuum robot prototype is the best choice to start our investigation of distributed control and distributed decision-making mechanisms. Continuum robots operate mainly in difficult-to-access areas, being wirelessly controlled. Thus, autonomy will be a very useful feature for continuum robots, especially in case of lost connection, when the robot can proceed with its work by itself using a decision-making mechanism. We provided a detailed description of the motion control models of continuum robots and revealed the problem of absence of models based on real physical robots' motions. For our further experiments, we realized a suitable tendon-driven continuum robot prototype and considered our steps for future work.

### REFERENCES

[1] O. Shamilyan, I. Kabin, Z. Dyka, M. Kuba, and P. Langendoerfer, "Octopuses: biological facts and technical solutions," in *2021 10th Mediterranean Conference on Embedded Computing (MECO)*, 2021, pp. 1–7.

[2] O. Shamilyan, I. Kabin, Z. Dyka, and P. Langendoerfer, "Distributed Artificial Intelligence as a Means to Achieve Self-X-Functions for Increasing Resilience: the First Steps," in *2022 11th Mediterranean Conference on Embedded Computing (MECO)*, 2022, pp. 1–6.

[3] K. Warwick, "Artificial intelligence." *The basics*. London, New York: Routledge, 2012.

[4] A. H. Bond and L. Gasser, "Readings in Distributed Artificial Intelligence." 1st ed. Burlington: Elsevier Science, 2014.

[5] G. Levy, N. Nesher, L. Zullo, and B. Hochner, "Motor Control in Soft-Bodied Animals," in *Oxford handbooks, The Oxford handbook of invertebrate neurobiology*, J. H. Byrne, Ed., New York, NY: Oxford University Press, 2017, pp. 494–510.

[6] P. Amodio, M. Boeckle, A. K. Schnell, L. Ostojíc, G. Fiorito, and N. S. Clayton, "Grow Smart and Die Young: Why Did Cephalopods Evolve Intelligence?," *Trends in ecology & evolution*, vol. 34, no. 1, pp. 45–56, 2019.

[7] S. Carls-Diamante, "Where Is It Like to Be an Octopus?," *Frontiers in systems neuroscience*, vol. 16, p. 840022, 2022.

[8] S. S. Adams and S. Burbeck, "Beyond the Octopus: From General Intelligence Toward a Human-Like Mind," in *Atlantis Thinking Machines, Theoretical Foundations of Artificial General Intelligence*, P. Wang and B. Goertzel, Eds., Paris: Scholars Portal, 2012, pp. 49–65.

[9] M. Jukic. "It's Time to Take Octopus Civilization Seriously." Palladium. [Online]. Available: https://www.palladiummag.com/2019/04/01/its-time-to-take-octopus-civilization-seriously Accessed on: Jan. 19, 2023.

[10] A. K. Schnell and N. S. Clayton, "Cephalopod cognition," *Current biology: CB*, vol. 29, no. 15, R726-R732, 2019

[11] T. Gutnick, R. A. Byrne, B. Hochner, and M. Kuba, "Octopus vulgaris uses visual information to determine the location of its arm," *Current biology: CB*, vol. 21, no. 6, pp. 460–462, 2011.
[12] J. E. Niven, "Invertebrate neurobiology: visual direction of arm movements in an octopus," *Current biology: CB*, vol. 21, no. 6, R217-8, 2011.
[13] C. Jozet-Alves, M. Bertin, and N. S. Clayton, "Evidence of episodic-like memory in cuttlefish," *Current biology: CB*, vol. 23, no. 23, R1033-5, 2013.
[14] G. Fiorito and P. Scotto, "Observational Learning in Octopus vulgaris," *Science (New York, N.Y.)*, vol. 256, no. 5056, pp. 545–547, 1992.
[15] "The Octopus: A Unique Animal for Studying the Brain." Frontiers for Young Minds. [Online]. Available: https://kids.frontiersin.org/articles/10.3389/frym.2021.752743 Accessed on: Oct. 17, 2022.
[16] T. Gutnick, M. J. Kuba, and A. Di Cosmo, "Neuroecology: Forces that shape the octopus brain," *Current biology: CB*, vol. 32, no. 3, R131-R135, 2022.
[17] A. K. Schnell, P. Amodio, M. Boeckle, and N. S. Clayton, "How intelligent is a cephalopod? Lessons from comparative cognition," *Biological reviews of the Cambridge Philosophical Society*, vol. 96, no. 1, pp. 162–178, 2021.
[18] R. E. Jacobs, "Diffusion MRI Connections in the Octopus Brain," *Experimental neurobiology*, vol. 31, no. 1, pp. 17–28, 2022.
[19] W.-S. Chung, N. D. Kurniawan, and N. J. Marshall, "Toward an MRI-Based Mesoscale Connectome of the Squid Brain," *iScience*, vol. 23, no. 1, p. 100816, 2020.
[20] T. Gutnick, T. Shomrat, J. A. Mather, and M. J. Kuba, "The Cephalopod Brain: Motion Control, Learning, and Cognition," in *Physiology of molluscs: A collection of selected reviews*, S. Saleuddin and S. Mukai, Eds., 1st ed., Toronto: Apple Academic Press, 2021, pp. 137–177.
[21] N. Nesher, F. Maiole, T. Shomrat, B. Hochner, and L. Zullo, "From synaptic input to muscle contraction: arm muscle cells of Octopus vulgaris show unique neuromuscular junction and excitation-contraction coupling properties," *Proceedings. Biological sciences*, vol. 286, no. 1909, p. 20191278, 2019.
[22] L. Zullo, H. Eichenstein, F. Maiole, and B. Hochner, "Motor control pathways in the nervous system of Octopus vulgaris arm," *J Comp Physiol A*, vol. 205, no. 2, pp. 271–279, 2019.
[23] T. Gutnick *et al.*, "Recording electrical activity from the brain of behaving octopus," *Current Biology*, vol. 33, no. 6, pp. 1171-1178.e4, Feb. 2023.
[24] Z. Y. Wang and C. W. Ragsdale, "Cephalopod Nervous System Organization," in *Oxford Research Encyclopedia of Neuroscience*, Z. Y. Wang and C. W. Ragsdale, Eds.: Oxford University Press, 2019.
[25] "Do octopuses' arms have a mind of their own? Researchers are unravelling the mystery of how octopuses move their arms." ScienceDaily. [Online]. Available: https://www.sciencedaily.com/releases/2020/11/201102120027.htm Accessed on: Jan. 19, 2023.
[26] S. Shigeno, P. L. R. Andrews, G. Ponte, and G. Fiorito, "Cephalopod Brains: An Overview of Current Knowledge to Facilitate Comparison With Vertebrates," *Frontiers in Physiology*, vol. 9, p. 952, 2018.
[27] B. Marenko, "FutureCrafting. A Speculative Method for an Imaginative AI," *AAAI Spring Symposia*, 2018. [Online]. Available: https://www.semanticscholar.org/paper/FutureCrafting.-A-Speculative-Method-for-an-AI-Marenko/960723aeeae5867c49cdb898d6647f1e20b42e8d
[28] A.-S. Al-Soudy, V. Maselli, S. Galdiero, M. J. Kuba, G. Polese, and A. Di Cosmo, "Identification and Characterization of a Rhodopsin Kinase Gene in the Suckers of Octopus vulgaris: Looking around Using Arms?," *Biology*, vol. 10, no. 9, 2021.
[29] A. Dorri, S. S. Kanhere, and R. Jurdak, "Multi-Agent Systems: A Survey," *IEEE Access*, vol. 6, pp. 28573–28593, 2018.
[30] Z. Brahmi, M. M. Gammoudi, H. Arioui, R. Merzouki, and H. A. Abbassi, "Decentralized method for complex task allocation in massive MAS," *Intelligent Systems and Automation*, vol. 1019, pp. 287–293, 2008.
[31] G. M. Skaltsis, H.-S. Shin, and A. Tsourdos, "A survey of task allocation techniques in MAS," in *2021 International Conference on Unmanned Aircraft Systems (ICUAS)*, Athens, Greece, 2021, pp. 488–497.
[32] A. Kantamneni, L. E. Brown, G. Parker, and W. W. Weaver, "Survey of multi-agent systems for microgrid control," *Engineering Applications of Artificial Intelligence*, vol. 45, pp. 192–203, 2015.
[33] S. Kim, C. Laschi, and B. Trimmer, "Soft robotics: a bioinspired evolution in robotics," *Trends in Biotechnology*, vol. 31, no. 5, pp. 287–294, 2013.
[34] P. E. Dupont, J. Lock, B. Itkowitz, and E. Butler, "Design and Control of Concentric-Tube Robots," *IEEE transactions on robotics : a publication of the IEEE Robotics and Automation Society*, vol. 26, no. 2, pp. 209–225, 2010.
[35] J. Burgner-Kahrs, D. C. Rucker, and H. Choset, "Continuum Robots for Medical Applications: A Survey," *IEEE transactions on robotics : a publication of the IEEE Robotics and Automation Society*, vol. 31, no. 6, pp. 1261–1280, 2015.
[36] J. Burgner-Kahrs, H. B. Gilbert, J. Granna, P. J. Swaney, and R. J. Webster, "Workspace characterization for concentric tube continuum robots," in *2014 IEEE/RSJ International Conference on Intelligent Robots and Systems*, Chicago, IL, USA, Sep. 2014 - Sep. 2014, pp. 1269–1275.
[37] J. Granna, A. Nabavi, and J. Burgner-Kahrs, "Computer-assisted planning for a concentric tube robotic system in neurosurgery," *Int J CARS*, vol. 14, no. 2, pp. 335–344, 2019.
[38] P. Rao, Q. Peyron, S. Lilge, and J. Burgner-Kahrs, "How to Model Tendon-Driven Continuum Robots and Benchmark Modelling Performance," *Frontiers in Robotics and AI*, vol. 7, p. 630245, 2020.
[39] A. Bajo, R. E. Goldman, L. Wang, D. Fowler, and N. Simaan, "Integration and preliminary evaluation of an Insertable Robotic Effectors Platform for Single Port Access Surgery," in *2012 IEEE International Conference on Robotics and Automation*, St Paul, MN, USA, 2012, pp. 3381–3387.
[40] J. Ding, R. E. Goldman, K. Xu, P. K. Allen, D. L. Fowler, and N. Simaan, "Design and Coordination Kinematics of an Insertable Robotic Effectors Platform for Single-Port Access Surgery," *IEEE/ASME transactions on mechatronics : a joint publication of the IEEE Industrial Electronics Society and the ASME Dynamic Systems and Control Division*, pp. 1612–1624, 2013.
[41] B. Ouyang, B. Liu, H.-Y. Tam, and D. Sun, "Design of an Interactive Control System for a Multisection Continuum Robot," *IEEE/ASME transactions on mechatronics : a joint publication of the IEEE Industrial Electronics Society and the ASME Dynamic Systems and Control Division*, vol. 23, no. 5, pp. 2379–2389, 2018.
[42] B. Mazzolai *et al.*, "Octopus-Inspired Soft Arm with Suction Cups for Enhanced Grasping Tasks in Confined Environments," *Advanced Intelligent Systems*, vol. 1, no. 6, p. 1900041, 2019.
[43] A. Gao, N. Liu, M. Shen, M. E M K Abdelaziz, B. Temelkuran, and G.-Z. Yang, "Laser-Profiled Continuum Robot with Integrated Tension Sensing for Simultaneous Shape and Tip Force Estimation," *Soft robotics*, vol. 7, no. 4, pp. 421–443, 2020.
[44] "Soft Robotics Area." place of pub. [Online]. Available: https://www.santannapisa.it/en/institute/biorobotics/soft-robotics-area Accessed on: Jan. 26, 22.
[45] J. B. Kahrs. "Continuum Robotics Laboratory." [Online]. Available: https://crl.utm.utoronto.ca/ Accessed on: Jan. 26, 22.
[46] "RoME Lab – Robotics and Medical Engineering (RoME) Lab." [Online]. Available: https://rome.cdm.depaul.edu/ Accessed on: Jan. 26, 22.
[47] "Octobot: A Soft, Autonomous Robot." Wyss Institute. [Online]. Available: https://wyss.harvard.edu/media-post/octobot-a-soft-autonomous-robot/ Accessed on: Feb. 2, 2022.
[48] T.-D. Nguyen and J. Burgner-Kahrs, "A tendon-driven continuum robot with extensible sections," in *2015 IEEE/RSJ International Conference on Intelligent Robots and Systems (IROS)*, Hamburg, Germany, Sep. 2015 - Oct. 2015, pp. 2130–2135.
[49] F. Renda, M. Giorelli, M. Calisti, M. Cianchetti, and C. Laschi, "Dynamic Model of a Multibending Soft Robot Arm Driven by Cables," *IEEE transactions on robotics : a publication of the IEEE Robotics and Automation Society*, vol. 30, no. 5, pp. 1109–1122, 2014.
[50] T. G. Thuruthel, "Machine Learning Approaches for Control of Soft Robots," Unpublished, 2019.

[51] I. S. Godage, D. T. Branson, E. Guglielmino, G. A. Medrano-Cerda, and D. G. Caldwell, "Dynamics for biomimetic continuum arms: A modal approach," in *IEEE-ROBIO 2011*, Karon Beach, Thailand, 2011, pp. 104–109.

[52] I. S. Godage, D. T. Branson, E. Guglielmino, and D. G. Caldwell, "Path planning for multisection continuum arms," in *2012 IEEE International Conference on Mechatronics and Automation*, Chengdu, China, 2012, pp. 1208–1213.

[53] Q. Peyron et al., "Magnetic concentric tube robots: Introduction and analysis," *The International Journal of Robotics Research*, vol. 41, no. 4, pp. 418–440, 2022.

[54] "ETH Zürich." [Online]. Available: https://ethz.ch/de.html Accessed on: Jun. 11, 2022.

[55] L. Burrows, "The first autonomous, entirely soft robot," *Harvard Gazette*, Aug, 2016. [Online]. Available: https://news.harvard.edu/gazette/story/2016/08/the-first-autonomous-entirely-soft-robot/ Accessed on: Jun. 22, 2022.

[56] M. Neumann and J. Burgner-Kahrs, "Considerations for follow-the-leader motion of extensible tendon-driven continuum robots," in *2016 IEEE International Conference on Robotics and Automation (ICRA 2016): Stockholm, Sweden, 16-21 May 2016*, Stockholm, 2016, pp. 917–923.

[57] Y. Xu, Q. Peyron, J. Kim, and J. Burgner-Kahrs, "Design of Lightweight and Extensible Tendon-Driven Continuum Robots using Origami Patterns," in *2021 IEEE 4th International Conference on Soft Robotics (RoboSoft)*, New Haven, CT, USA, 2021, pp. 308–314.

[58] K. Nuelle, T. Sterneck, S. Lilge, D. Xiong, J. Burgner-Kahrs, and T. Ortmaier, "Modeling, Calibration, and Evaluation of a Tendon-Actuated Planar Parallel Continuum Robot," *IEEE Robot. Autom. Lett.*, vol. 5, no. 4, pp. 5811–5818, 2020.

[59] F. Renda, F. Boyer, J. Dias, and L. Seneviratne, "Discrete Cosserat Approach for Multisection Soft Manipulator Dynamics," *IEEE transactions on robotics : a publication of the IEEE Robotics and Automation Society*, vol. 34, no. 6, pp. 1518–1533, 2018.

[60] F. Renda, C. Armanini, V. Lebastard, F. Candelier, and F. Boyer, "A Geometric Variable-Strain Approach for Static Modeling of Soft Manipulators With Tendon and Fluidic Actuation," *IEEE Robot. Autom. Lett.*, vol. 5, no. 3, pp. 4006–4013, 2020.

[61] H. Alfalahi, F. Renda, and C. Stefanini, "Concentric Tube Robots for Minimally Invasive Surgery: Current Applications and Future Opportunities," *IEEE Trans. Med. Robot. Bionics*, vol. 2, no. 3, pp. 410–424, 2020.

[62] T. G. Thuruthel, E. Falotico, F. Renda, and C. Laschi, "Learning dynamic models for open loop predictive control of soft robotic manipulators," *Bioinspiration & biomimetics*, vol. 12, no. 6, p. 66003, 2017.

[63] T. G. Thuruthel, E. Falotico, F. Renda, T. Flash, and C. Laschi, "Emergence of behavior through morphology: a case study on an octopus inspired manipulator," *Bioinspiration & biomimetics*, vol. 14, no. 3, p. 34001, 2019.

[64] T. G. Thuruthel, E. Falotico, F. Renda, and C. Laschi, "Model-Based Reinforcement Learning for Closed-Loop Dynamic Control of Soft Robotic Manipulators," *IEEE transactions on robotics : a publication of the IEEE Robotics and Automation Society*, vol. 35, no. 1, pp. 124–134, 2019.

[65] T. George Thuruthel, Y. Ansari, E. Falotico, and C. Laschi, "Control Strategies for Soft Robotic Manipulators: A Survey," *Soft robotics*, vol. 5, no. 2, pp. 149–163, 2018.

[66] M. Cianchetti, A. Arienti, M. Follador, B. Mazzolai, P. Dario, and C. Laschi, "Design concept and validation of a robotic arm inspired by the octopus," *Materials Science and Engineering: C*, vol. 31, no. 6, pp. 1230–1239, 2011.

[67] "Soft Continuum Manipulator Research." RoME Lab. [Online]. Available: https://rome.cdm.depaul.edu/soft-continuum-manipulator-research/ Accessed on: Jul. 23, 2022.

[68] "Soft Continuum Snake Robots – RoME Lab." [Online]. Available: https://rome.cdm.depaul.edu/soft-continuum-snake-robots/ Accessed on: Jul. 23, 2022.

[69] "Concentric tube robotic tools for Frontal Sinus surgery." RoME Lab. [Online]. Available: https://rome.cdm.depaul.edu/concentric-tube-robotic-tools-for-frontal-sinus-surgery/ Accessed on: Jul. 23, 2022.

[70] "Minimally Invasive Intervention for Intracerebral Hemorrhage." RoME Lab. [Online]. Available: https://rome.cdm.depaul.edu/minimally-invasive-intervention-for-intracerebral-hemorrhage/ Accessed on: Jul. 23, 2022.

[71] L. Fichera et al., "Through the Eustachian Tube and Beyond: A New Miniature Robotic Endoscope to See Into The Middle Ear," *IEEE robotics and automation letters*, vol. 2, no. 3, pp. 1488–1494, 2017.

[72] I. S. Godage, E. Guglielmino, D. T. Branson, G. A. Medrano-Cerda, and D. G. Caldwell, "Novel modal approach for kinematics of multisection continuum arms," in *2011 IEEE/RSJ International Conference on Intelligent Robots and Systems*, San Francisco, CA, 2011, pp. 1093–1098.

[73] C. Della Santina, A. Bicchi, and D. Rus, "On an Improved State Parametrization for Soft Robots With Piecewise Constant Curvature and Its Use in Model Based Control," *IEEE Robot. Autom. Lett.*, vol. 5, no. 2, pp. 1001–1008, 2020.

[74] G. Runge, M. Wiese, L. Gunther, and A. Raatz, "A framework for the kinematic modeling of soft material robots combining finite element analysis and piecewise constant curvature kinematics," in *2017 3rd International Conference on Control, Automation and Robotics (ICCAR)*, Nagoya, Japan, 2017, pp. 7–14.

[75] M. W. Hannan and I. D. Walker, "Kinematics and the implementation of an elephant's trunk manipulator and other continuum style robots," *Journal of robotic systems*, vol. 20, no. 2, pp. 45–63, 2003.

[76] R. K. Katzschmann, C. Della Santina, Y. Toshimitsu, A. Bicchi, and D. Rus, "Dynamic Motion Control of Multi-Segment Soft Robots Using Piecewise Constant Curvature Matched with an Augmented Rigid Body Model," in *RoboSoft 2019: 2019 IEEE International Conference on Soft Robotics : April 14-18, 2019, COEX, Seoul, Korea, Seoul, Korea (South)*, 2019, pp. 454–461.

[77] H. Cheng, H. Liu, X. Wang, and B. Liang, "Approximate Piecewise Constant Curvature Equivalent Model and Their Application to Continuum Robot Configuration Estimation," in *2020 IEEE International Conference on Systems, Man, and Cybernetics (SMC)*, Toronto, ON, Canada, 2020, pp. 1929–1936.

[78] S. Kolachalama and S. Lakshmanan, "Continuum Robots for Manipulation Applications: A Survey," *Journal of Robotics*, vol. 2020, pp. 1–19, 2020.

[79] "Théorie des Corps déformables," *Nature*, vol. 81, no. 2072, p. 67, 1909.

[80] "Cosserat Rods." Univeristy of Illinois, Gazzola Lab. [Online]. Available: https://www.cosseratrods.org/ Accessed on: 18-Aug-22.

[81] N. Nesher and T. Shomrat, "How Does the Octopus Efficiently Control Its Flexible, Multi-Armed Body?," *Front. Young Minds*, vol. 9, 2021.

[82] J. N. Richter, B. Hochner, and M. J. Kuba, "Octopus arm movements under constrained conditions: adaptation, modification and plasticity of motor primitives," *The Journal of experimental biology*, vol. 218, Pt 7, pp. 1069–1076, 2015.

[83] D. C. Rucker, B. A. Jones, and R. J. Webster, "A Geometrically Exact Model for Externally Loaded Concentric-Tube Continuum Robots," *IEEE transactions on robotics : a publication of the IEEE Robotics and Automation Society*, vol. 26, no. 5, pp. 769–780, 2010.

[84] D. C. Rucker and R. J. Webster III, "Statics and Dynamics of Continuum Robots With General Tendon Routing and External Loading," *IEEE transactions on robotics : a publication of the IEEE Robotics and Automation Society*, vol. 27, no. 6, pp. 1033–1044, 2011.

[85] R. J. Webster and B. A. Jones, "Design and Kinematic Modeling of Constant Curvature Continuum Robots: A Review," *The International Journal of Robotics Research*, vol. 29, no. 13, pp. 1661–1683, 2010.

[86] D. B. Camarillo, C. R. Carlson, and J. K. Salisbury, "Configuration Tracking for Continuum Manipulators With Coupled Tendon Drive," *IEEE transactions on robotics : a publication of the IEEE Robotics and Automation Society*, vol. 25, no. 4, pp. 798–808, 2009.

[87] J. M. P. Menezes and G. A. Barreto, "Long-term time series prediction with the NARX network: An empirical evaluation," *Neurocomputing*, vol. 71, 16-18, pp. 3335–3343, 2008.

[88] E. Diaconescu, "The use of NARX neural networks to predict chaotic time series," *WSEAS Transactions on Computer Research*, vol. 3, no. 3, pp. 182–191, 2008.

[89] I. S. Godage, D. T. Branson, E. Guglielmino, G. A. Medrano-Cerda, and D. G. Caldwell, "Shape function-based kinematics and dynamics for variable length continuum robotic arms," in *2011 IEEE International Conference on Robotics and Automation (ICRA)*, Shanghai, China, 2011, pp. 452–457.

[90] "Changing Your Idea of What Robots Can Do." Boston Dynamics. [Online]. Available: https://www.bostondynamics.com/ Accessed on: Sep. 24, 2022.

[91] "Boston Dynamics leverages cognitive and athletic intelligence to create human-like robots." AUVSI. [Online]. Available: https://www.auvsi.org/industry-news/boston-dynamics-leverages-cognitive-and-athletic-intelligence-create-human-robots Accessed on: Oct. 7, 2022.

[92] "Spot® - The Agile Mobile Robot." Boston Dynamics. [Online]. Available: https://www.bostondynamics.com/products/spot Accessed on: Aug. 11, 2022.

[93] "octopus_dynamics." GitHub. [Online]. Available: https://github.com/tomraven1/octopus_dynamics Accessed on: Aug. 20, 2022.

[94] S. Zimmermann, R. Poranne, and S. Coros, "Go Fetch! - Dynamic Grasps using Boston Dynamics Spot with External Robotic Arm," in *2021 IEEE International Conference on Robotics and Automation (ICRA)*, pp. 4488–4494.

[95] "Soft Robotics Toolkit." [Online]. Available: https://softroboticstoolkit.com/ Accessed on: Sep. 17, 2022.

[96] J. Vasquez. "The Bootup Guide To Homebrew Two-Stage Tentacle Mechanisms." Hackaday. [Online]. Available: https://hackaday.com/2016/09/13/the-bootup-guide-to-homebrew-two-stage-tentacle-mechanisms/ Accessed on: Jan. 21. 2022.

[97] J. Vasquez. "Two-Stage Tentacle Mechanisms Part II: The Cable Controller." Hackaday. [Online]. Available: https://hackaday.com/2016/10/05/two-stage-tentacle-mechanisms-part-ii-the-cable-controller/ Accessed on: Jan. 21. 2022.

[98] J. Vasquez. "Two-Stage Tentacle Mechanisms Part III: Putting It All Together." Hackaday. [Online]. Available: https://hackaday.com/2016/10/21/two-stage-tentacle-mechanisms-part-iii-putting-it-all-together/ Accessed on: Jan. 21. 2022.

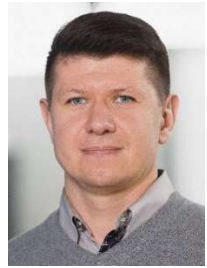

**IEVGEN KABIN** received the Diploma degree in Electronic Systems from the National Technical University of Ukraine ''Kyiv Polytechnic Institute'', Kiev, Ukraine, in 2009. From 2009 to 2010, he was a Leading Engineer with the state-owned enterprise, Scientific Production Center of Energy-efficient Designs and Technologies, ("Tehnoluch"). From 2010 to 2015, he was a Junior Researcher with the E.O. Paton Electric Welding Institute, National Academy of Sciences of Ukraine. Since 2015, he is with IHP-Leibniz Institut für innovative Mikroelektronik, Frankfurt (Oder), Germany. There he is a member of the research group Sensor Networks and Middleware Platforms Group and published more than 15 articles. His research interests include side-channel analysis attacks and investigations in the field of Elliptic Curve Cryptography.

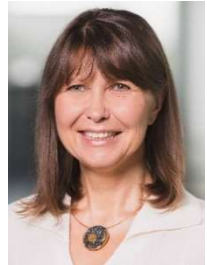

**ZOYA DYKA** received her Diploma degree in Radiophysics and Electronics from Taras Shevchenko University Kiev, Ukraine in 1996 and the Ph.D. degree from Technical University of Cottbus-Senftenberg, Germany in 2012. Since 2000 she is with the IHP in Frankfurt (Oder). Since 2013, she has been leading a young researchers group in the field of tamper resistant crypto ICs. Since 2018, she is leading a research group in the field of resilient CPSoS. She has authored more than 30 peer reviewed technical articles and filed five patents in the security area of which four are granted. Her research interests include design of efficient hardware accelerators for cryptographic operations, SCA countermeasures, anti-tampering means and resilience.

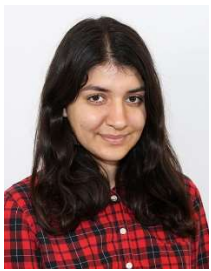

**OXANA SHAMILYAN** received the Diploma degree in Information Security from the Southern Federal University, Taganrog, Russia, in 2020. Since 2020 she is with IHP-Leibniz Institut für innovative Mikroelektronik, Frankfurt (Oder), Germany. There she is a member of Total Resilience group and doing her Ph.D. thesis in a field of distributed AI approaches. Her research interests include decision-making and investigation of distributed AI approaches.

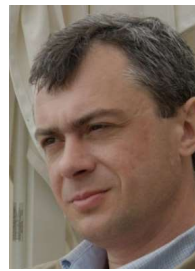

**OLEKSANDR SUDAKOV** Candidate of physical and mathematical sciencies (Ph.D.). Associate Professor of Medical Radiophysics Department, Faculty of Radiophysics, Electronics and Computer Systems, Head of Parallel Computing Laboratory at Information and Computer Center Taras Shevchenko National University of Kyiv. Graduated from Radiophysics Faculty Taras Shevchenko Kyiv University in 1996. Ph.D. thesis in Procession of magnetic resonance tomography signals in 2002. Scientific interests: high performance computing, physical processes in biological systems.

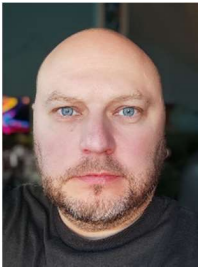

**ANDRII CHERNINSKYI** PhD in biology, senior researcher at Bogomoletz Institute of Physiology, National Academy of Sciences of Ukraine. He studies the electrophysiological correlates of behavioral changes in animals induced by alteration of cellular functions in neurons. The research interests also include brain functioning, EEG, ERP, ICA, human psychophysiology, animal's behavior, emotions, odor's perception, Parkinson's disease, acid-sensing ion channels, and epilepsy.

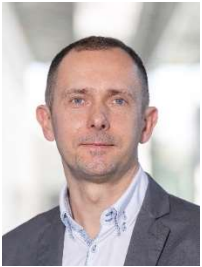

**MARCIN BRZOZOWSKI** received his diploma and Ph.D. degrees in computer science from BTU Cottbus, Germany in 2006 and 2012 respectively. Since 2006 he is with the IHP in Frankfurt (Oder) and has mainly worked on embedded systems, sensor networks, and adaptive networking. Since 2022 he leads the research group Elastic Computing.

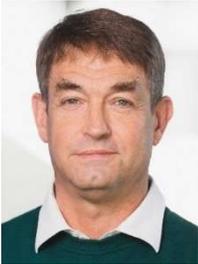

**PROF. DR. PETER LANGENDÖRFER** holds a diploma and a doctorate degree in computer science. Since 2000 he is with the IHP in Frankfurt (Oder). There, he is leading the wireless systems department. From 2012 till 2020 he was leading the chair for security in pervasive systems at the Technical University of Cottbus-Senftenberg. Since 2020 he owns the chair wireless systems at the Technical University of Cottbus-Senftenberg. He has published more than 145 refereed technical articles, filed 17 patents of which 10 have been granted already. He worked as guest editor for many renowned journals e.g. Wireless Communications and Mobile Computing (Wiley) and ACM Transactions on Internet Technology. Peter is highly interested in security for resource constraint devices, low power protocols and resilience.